\documentclass{article}
\usepackage{microtype}
\usepackage{graphicx}
\usepackage{booktabs} 
\usepackage{multirow}
\usepackage{amsmath}
\usepackage{amssymb}
\usepackage{mathtools}
\usepackage{amsthm}
\usepackage{wrapfig}
\usepackage[most]{tcolorbox}
\usepackage{colortbl}
\usepackage{xcolor}
\usepackage{tabularx}
\usepackage{enumitem}

\usepackage{caption}
\usepackage{newtxmath}
\usepackage{subcaption}
\usepackage{algorithm}
\usepackage[endLComment=,italicComments=false]{algpseudocodex}
\usepackage{hyperref}

\def\mathcolor#1#{\@mathcolor{#1}}
\def\@mathcolor#1#2#3{%
  \protect\leavevmode
  \begingroup
    \color#1{#2}#3%
  \endgroup
}

\newcommand{\wrapfill}{\par\ifnum\value{WF@wrappedlines}>0
  \addtocounter{WF@wrappedlines}{-1}%
  \null\vspace{\arabic{WF@wrappedlines}\baselineskip}%
  \WFclear
\fi}

\definecolor{lightgreen}{HTML}{CEEAD6}
\definecolor{lightred}{HTML}{FAD2CF}
\definecolor{lightorange}{HTML}{FEEFC3}
\definecolor{lightblue}{HTML}{30CEFE}
\definecolor{darkerblue}{HTML}{5CA3FF}
\definecolor{brightpurple}{HTML}{9865FE}
\definecolor{dimblue}{HTML}{285CAA}

\newtcolorbox[auto counter, number within=section]{remarkbox}[1][]{
    colback=olive!5!white,   
    colframe=olive!30!black,   
    fonttitle=\small\itshape, 
    title=Remark~\thetcbcounter: #1,
    boxrule=0.5pt,            
    arc=3mm                   
}
\usepackage[preprint]{corl_2026} 

\usepackage[capitalize,noabbrev]{cleveref}

\renewcommand{\eqref}[1]{(\ref{#1})}
\newcommand{\eg}{\emph{e.g.}}
\newcommand{\ie}{\emph{i.e.}}

\DeclareMathOperator*{\argmin}{arg\,min}

\makeatletter
\Crefname{section}{\S\@gobble}{\S\@gobble}
\Crefname{subsection}{\S\@gobble}{\S\@gobble}
\Crefname{proposition}{Prop.}{Props.}
\Crefname{figure}{Fig.}{Figs.}

\renewcommand{\paragraph}{%
  \@startsection{paragraph}{4}{\z@}
                {0.ex}
                {-1em}
                {\normalsize\bf}
}

\makeatother

\title{Optimal Transport Q-Learning \\for Flow Policy Steering and Acceleration}

\author{
  Andreas Sochopoulos$^{1,2}$,\ Esmeralda S. Whitammer$^1$, \ Nikolaos Tsagkas$^1$,\ Jo\~ao Moura$^1$, \\ 
  \textbf{Michael Gienger$^2$,\ Sethu Vijayakumar$^1$} \\
  $^1$University of Edinburgh \ $^2$Honda Research Institute Europe\\
  \texttt{\href{https://ansocho.github.io/otql-flow/}{ansocho.github.io/otql-flow}}\\
}

\begin{document}
\maketitle


\begin{abstract}
     Diffusion and flow policies have recently demonstrated remarkable performance in robotic applications by accurately capturing multimodal robot trajectory distributions, especially in the context of vision language action (VLA) models.
     However, high quality policy performance also requires fast inference and high quality demonstrations, which are often hard to get. 
     Lack of these leads to suboptimal policy behaviors and failure under distribution shifts. 
     In this work we address the problem of fine-tuning and accelerating suboptimal flow-based policies using the robot's experience through RL post-training. 
     We introduce \textit{Optimal Transport Q-Learning} (OTQL), a new methodology for finetuning flow policies using advantage weighted conditional optimal transport flow matching. 
     OTQL can finetune and accelerate flows with an interaction budget of 50-60 episodes while avoiding computationally expensive distillation in simulation and real-world robot tasks. 
     Our results show that OTQL post-trains flow policies using the robot's own experience, increasing average success percentage of single-task policies from $36\%$ to $86\%$ and of a pre-trained VLA from $38\%$ to $76\%$ while reducing the number of inference steps per action generation by $70\%$.
\end{abstract}

\keywords{Flow Matching, Reinforcement Learning, Optimal Transport} 


\section{Introduction}
\vspace{-6pt}
\label{sec:intro}
	
Vision language action models (VLAs) and, more recently, world action models (WAMs) have brought the field of robotics one step closer to generalist policies. 
These systems use as backbone a large, pre-trained vision language model (VLM) \citep{yang2025qwen3, beyer2024paligemma} or world model (WM), often in the form of a video model~\citep{pai2025mimic, kim2026cosmos}, that provides features to an action head. 
This module is most commonly a flow or diffusion model \citep{kim2026cosmos, pai2025mimic, reuss2025flower, pmlr-v305-black25a, BlackK-RSS-25, bjorck2025gr00t}. 
Flows trained with flow matching (FM) \citep{lipman2022flow, tong2023improving} and diffusion models \citep{ho2020denoising, song2020score} have proven highly effective for action generation, as they can capture complex, multimodal action distributions \citep{chi2023diffusion, pmlr-v162-janner22a}, such as those demonstrated by humans during teleoperation or kinesthetic teaching. 

Despite undergoing large-scale pretraining, VLAs and WAMs often struggle with zero-shot generalization to tasks unseen during training. 
Moreover, single and multi-task flow and diffusion policies frequently fail to handle out-of-distribution environment states, leading to task failures. 
The standard remedy for these issues is collecting additional demonstrations for the target tasks or states, and performing supervised fine-tuning (SFT) on the newly acquired dataset. 
However, collecting data through teleoperation is tedious and requires significant human effort, time, and expertise.

Reinforcement learning (RL) provides the tools for policies to adapt using the robot's own experience, eliminating the need for human demonstrations \citep{pmlr-v305-zhang25a, pmlr-v305-wagenmaker25a, du2026dynaguide, ankile2025imitation, ankile2025residual}. 
A large body of work has attempted to train flow and diffusion policies in \textit{offline RL} settings using a static set of rollouts \citep{wang2023diffusion, tiofack2025guided, zhang2025energyweighted, li2026reinforcement, li2026q, park2025flow}. 
Although successful in simple to moderate robotic tasks, most of these methods suffer from \textit{slow inference} because they require integrating an SDE or ODE over multiple steps \citep{li2026reinforcement}. 
Furthermore, since critic training typically involves generating actions from the learned policy, this slow inference propagates to every training step, resulting in \textit{slow training}. 
While there have been attempts to train single-step policies \citep{park2025flow, tiofack2025guided}, they rely on computationally expensive distillation. 
In the context of behavior cloning, there has been extensive 
work on diffusion/flow acceleration \citep{wang2024one, prasad2024consistency, sochopoulos2025fast, hu2024adaflow, lu2024manicm, ding2024fast}, but methods that address both acceleration and RL post-training remain scarce.

In this paper, we present \textit{Optimal Transport Q-Learning (OTQL)}, an FM methodology that \textit{steers} flow-based policies to achieve \textit{few-step inference} without expensive training. 
Our method uses conditional optimal transport (COT) couplings \citep{sochopoulos2025fast, kerrigan2024dynamic, Cheng_2025_ICCV} between noise and action samples to train flows with straight integration paths, avoiding the computational complexity of common inference acceleration techniques \citep{yin2024one, salimans2022progressive, frans2024one, park2025flow, kim2023consistency, yang2024consistency}. 
Furthermore, we train a critic function and use it to inform the marginal probabilities of the action distribution while solving the COT problem. 
This results in couplings that favor high-value samples during training, leading to flows that learn straight paths between the noise distribution and these optimal samples.

Flows steered with OTQL demonstrate strong empirical performance in both offline and online fine-tuning scenarios, requiring only two to three integration steps for inference. 
More specifically, we show that OTQL achieves fine-tuned performance comparable to state-of-the-art steering methods \citep{pmlr-v305-wagenmaker25a, li2026reinforcement, zhang2025energyweighted} in both real and simulated tasks, but at a fraction of the training and inference time. 
We also demonstrate OTQL's ability to steer VLAs (\eg, SmolVLA \citep{shukor2025smolvla}) on novel tasks using the robot's own experience. 
In real-world tasks, our approach increases SmolVLA's performance from $38\%$ to $76\%$ while reducing the neural function evaluations (NFEs) required for inference by $70\%$.
	
\section{Background}
\vspace{-6pt}
\label{sec:background}

\paragraph{Offline and online RL.} We consider a Markov decision process $\mathcal{M} = (\mathcal{S}, \mathcal{A}, r, \rho_0, \gamma, P)$, where $\mathcal{S}$ is the state space, $\mathcal{A} \in \mathbb{R}^d$ the action space, $r(s_t, a_t)$ the reward function, $\rho_0(s_0) \in \Delta(\mathcal{S})$ the initial state distribution, $\gamma \in [0, 1)$ the discount factor, and $P(s_{t+1} \mid s_t, a_t)$ the transition dynamics. By $\Delta(\mathcal{X})$ we denote the set of probability distributions defined over the set $\mathcal{X}$. Assuming a policy $\pi(a\mid s):\mathcal{S}\rightarrow\Delta(\mathcal{A})$ that maps states to actions we define the critic over $\pi$ as $Q_\pi(s_t,a_t) = \mathbb{E}_\pi\left[\sum_{t=0}^\infty \gamma^t r(s_t, a_t) \mid s_0=s_t, a_0=a_t\right]$. The goal of online RL is to find a policy $\pi$ that maximizes the expected discounted return $\mathbb{E}_{s_{t+1}\sim P(\cdot\mid s_t, a_t), a_t\sim\pi(\cdot\mid s_t)}\left[\sum_{t=0}^\infty \gamma^t r(s_t, a_t)\right]$ with environment interactions, while the goal of offline RL is to find a policy that maximizes the expected discounted return using only an offline dataset $\mathcal{D}$ of transitions $(s, a, r, s')$.

\paragraph{Imitation learning as conditional generative modeling.}
In imitation learning (IL), we typically assume access to a dataset of expert demonstrations $\mathcal{D}=\{(s^{(i)},a^{(i)})\}_{i=1}^n$ , which contains observation-action sequence pairs. 
Given the dataset $\mathcal{D}$ we treat the learning of actions $a$ as a conditional generative modeling problem conditioned on $s$. We treat the $a^{(i)}$ as samples from a target distribution $p_a(a\mid s)$ with conditions $s=s^{(i)}$. 
In the context of generative modeling with continuous normalizing flows, or neural ODEs \citep{chen2018neural}, we are searching for an ODE that transforms an initial noise distribution $p_z$ over $\mathbb{R}^d$ into the conditional action distribution $p_a(a\mid s)$. Given an ODE $dx = v_\theta(t, x\mid s)\,dt$ , 
where $v_\theta:\mathbb{R}\times\mathbb{R}^d\times\mathcal{S}\to\mathbb{R}^d$ is a vector field parametrized by a neural network with parameters $\theta$ taking $x, s$ and $t$ as input, one aims to find $\theta$ such that if $x(0)=z\sim p_z$, then the distribution over $x(1)$ induced by integration of the ODE from $t=0$ to $t=1$ with initial conditions $x(0)$ matches $p_a$. 

\paragraph{RL post-training.} Assuming we are given a policy that samples from a conditional distribution $\pi(a\mid s)$ approximating a distribution $p_a(a\mid s)$, we are interested in sampling from $\tilde{\pi}(a \mid s) \propto \pi(a \mid s) e^{ \lambda Q(s,a)}$ or equivalently $\tilde{\pi}(a \mid s) \propto \pi(a \mid s) e^{ \lambda A(s,a)}$, where $A$ is the advantage function $A(s_t,a_t) = Q(s_t,a_t) - V(s_t)$ and $V(s_t) = \mathbb{E}_{s_{t+1}\sim P(s_t, a_t), a_t \sim \pi}[Q(s_t,a_t)]$ is the state-value function. Our motivation for attempting to sample $\tilde{\pi}$ is grounded on the fact that it is the closed-form solution to the KL regularized problem as defined in \citep{peters2007reinforcement, peters2010relative, nair2020awac}. Intuitively, we are trying to \textit{steer} the base policy towards behaviors that it already models and that maximize the $Q$ value. 

 \paragraph{Conditional flow matching.} In this paper we focus on policies instantiated as neural ODE-based generative models which are trained with CFM. To implement CFM, one must specify a \emph{coupling} distribution $q(z, a\mid s)$ whose marginals equal $p_z$ and $p_a$ and an interpolating trajectory that links each pair $(z, a)$. This trajectory is governed by an ordinary differential equation (ODE), $dx=u(t,x\mid z,a)\,dt$, constrained such that the state flows from $x(0)=z$ to $x(1)=a$. Throughout this work, we restrict our focus to a linear interpolant, $u(t,x\mid z,a)=a-z$. Consequently, the state evolves as $x(t)=ta+(1-t)z,\; t \in [0,1]$.  Under this linear formulation, the neural network $v_\theta$ is optimized via the following stochastic regression loss:
\begin{align}
    \label{eq:cfm_loss}
    \mathcal{L}_{\text{CFM}}(\theta) := \mathop{\mathbb{E}}_{\substack{t \sim \mathcal{U}([0,1]) \\ z,a\sim q(z,a\mid s)}} \big\| \underbrace{v_\theta(t, \overbrace{ta+(1-t)z}^{\text{interpolant}} \mid s)}_{\text{learned vector field}} - \underbrace{(a-z)}_{\text{target}} \big\|^2.
\end{align}
The primary theoretical justification for this objective is that the optimal time-dependent vector field $v_\theta$ minimizing \eqref{eq:cfm_loss} successfully pushes the base distribution $p_z$ to the target distribution $p_a$ at $t=1$. This effectively resolves the generative modeling task defined earlier.

\paragraph{Conditional optimal transport flow matching.} Different choices exist for the coupling $q(z,a)$. In Independent Coupling CFM $q(z,a)=q(z)q(a)$. Another option, which was shown in \cite{tong2023improving,pooladian2023multisample} to reduce objective variance and straighten integration curves in the unconditional case, takes $q$ to be a minibatch OT plan computed from batches of samples $z,a$. Such a coupling approximates the OT plan between $p_z$ and $p_a$. A similar idea has also been explored in the case of conditional distributions \citep{kerrigan2024dynamic, sochopoulos2025fast, Cheng_2025_ICCV} using minibatch OT to approximate a conditional OT (COT) plan. For more details on the basic definitions of OT and COT see \citep{tong2023improving} and \Cref{sec:appendix_ot}. 

Describing the OT plan as the solution of an ODE that moves particles from $z$ to $a$ over time yields the dynamic formulation of the OT problem, commonly known as the Benamou--Brenier formulation~\citep{benamou2000computational}.
The dynamic OT plan transports particles along straight-line trajectories from $z$ to $a$ and when $q(z, a)$ denotes the 2-Wasserstein OT coupling (see \Cref{sec:appendix_ot})
, the vector field that minimizes the unconditional version of the CFM objective in \eqref{eq:cfm_loss} coincides exactly with the dynamic OT ODE \citep{tong2023improving}. For a more detailed discussion, see \citet{tong2023improving}. \citet{kerrigan2024dynamic} have extended these results to the dynamic formulation of COT.


\section{Optimal Transport Q-Learning}
\vspace{-6pt}
\label{sec:otql}

In this section we present OTQL, our method for learning flow policies that are able to sample actions from $\tilde{\pi}(a\mid s)$. In \Cref{sec:otql_cotcfm} we present the CFM training loss that enables OTQL and in \Cref{sec:otql_rl} we detail the full algorithm.

\subsection{Conditional optimal transport for flow acceleration and steering}
\label{sec:otql_cotcfm}
Motivated by the observations made in \Cref{sec:background} connecting OT and CFM, our goal is to learn a flow that approximates the dynamic COT ODE between the noise distribution and $\tilde{\pi}$. We initially consider a general version of the form $\tilde{\pi}(a\mid c)\propto\pi(a\mid c)e^{-\lambda\mathcal{E}(a, c)}$, where c is a general condition vector and $\mathcal{E}:\mathbb{R}^d\times\mathbb{R}^d\rightarrow\mathbb{R}$ is a positive energy function. To learn a flow that samples from $\tilde{\pi}$, we employ an approximate COT plan as the coupling $q(z, \tilde{a} \mid c)$, where $\tilde{a}\sim \tilde{\pi}$, between the two distributions. The resulting flow is intended to approximate the COT ODE linking the noise distribution and samples from $\tilde{\pi}$, thereby producing straighter trajectories towards low-energy samples. 

\begin{figure}[t]
\centering
\includegraphics[width=\linewidth]{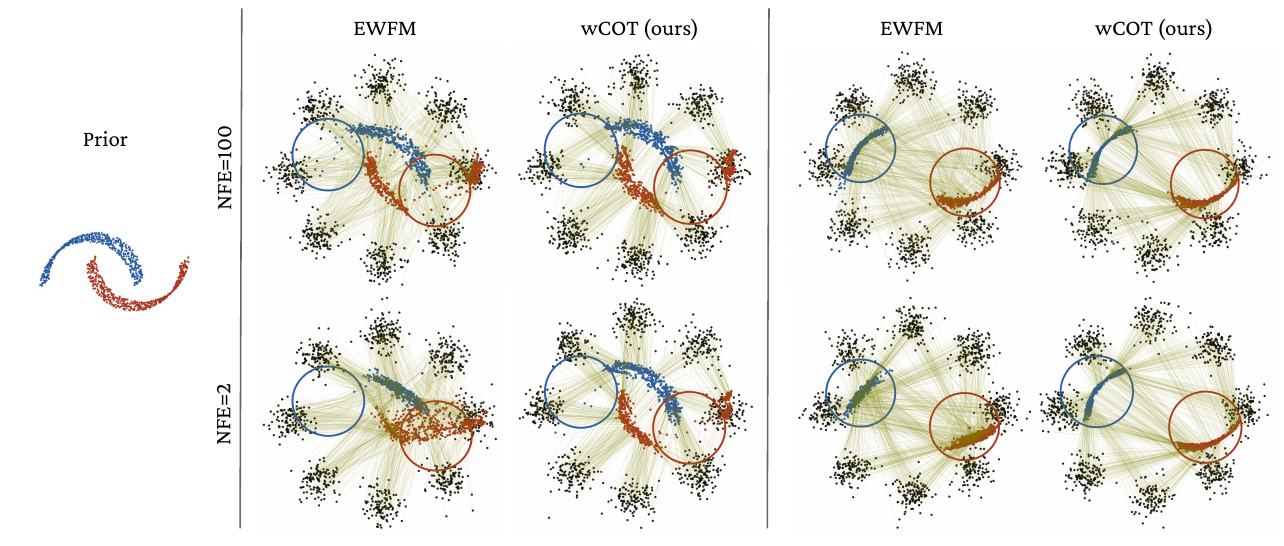}
\caption{Energy-weighted Flow Matching (EWFM) \cite{zhang2025energyweighted} and weighted-marginal COT flow matching (wCOT) flows trained to generate the tilted conditional two moons distribution from the 8 Gaussians distribution. Generation with 100 (top row) and 2 (bottom row) \texttt{euler} integration steps is shown. The samples generated from both methods given an energy function that is $1$ inside the drawn circles and $0$ everywhere else is shown in the \textbf{left} and given an energy function that is $1$ outside and $0$ inside the circles in the \textbf{right}.}
\vspace*{-1em}
\label{fig:moons}
\end{figure}

\paragraph{Unconditional OT couplings}
In practice, given $m$ samples from the dataset $\{(a_j, c_j)\}_{j=1}^m \sim \mathcal{D}$, where $c_j$ are conditions, and $m$ noise samples $\{z_i\}_{i=1}^m \sim \mathcal{N}(0, I_d)$ we compute a mini-batch approximation of the unconditional OT plan by
solving the following optimization problem between the measures $\mathbf{z} = \sum_{i=0}^mk_i\delta_{z_i}$ and $\mathbf{a} = \sum_{j=0}^mp_i\delta_{a_j}$ (where typically $k_i=p_i=\frac{1}{m}$):
\begin{equation}
    T_\varepsilon(\mathbf{z}, \mathbf{a})
    \;=\;
    \min_{\Gamma \,\in\, \Gamma(\mathbf{z},\,\mathbf{a})}
    \sum_{i,j} C_{ij}\,\Gamma_{ij}
    -
    \varepsilon\,H(\Gamma), \quad s.t.\; \sum_j\Gamma_{ij}=k_i,\; \sum_i\Gamma_{ij}=p_j
    \label{eq:eot}
\end{equation}

where $C_{ij} = \|x_i - x_j\|^2$ is the squared-Euclidean cost,
$H(\Gamma) = -\sum_{i,j}\Gamma_{ij}\log(\Gamma_{ij}-1)$ is the entropy
regulariser, $\varepsilon > 0$ is the regularisation strength, and
$\Gamma(\mathbf{z}, \mathbf{a})$ is the set of joint distributions
(transport plans) with marginals $\mathbf{z}$ and $\mathbf{a}$.

\paragraph{Conditioning via cost augmentation.}
Since we target \emph{conditional} action distributions (\ie action distributions conditioned on robot observations),
we also wish the coupling to respect the condition structure: source
and target points with the same condition should be preferentially
paired, so the triangular constraint of COT \citep{kerrigan2024dynamic} is satisfied. Similar to \citep{sochopoulos2025fast}, we incorporate the
condition directly into the cost by concatenating it with the data
vector before solving \eqref{eq:eot}. Concretely, we form augmented vectors
\begin{equation}
    \bar{z}_i = \bigl[z_i \mathcolor{dimblue}{\;\|\; \alpha \bar{c}_i}\bigr],
    \quad
    \bar{a}_j = \bigl[a_j \mathcolor{dimblue}{\;\|\; \alpha c_j}\bigr],
    \label{eq:concat}
\end{equation}
where $\alpha > 0$ is a scaling hyperparameter controlling the relative
influence of the condition on the coupling, and $\bar{c_i}$ are the elements of the source condition vector, the choice of which will be explained in the next paragraph. The augmented cost
becomes:
\begin{equation}
    C_{ij}
    =
    \bigl\|\bar{z}_i - \bar{a}_j\bigr\|^2
    =
    \|z_i - a_j\|^2
    \mathcolor{dimblue}{+\alpha^2\,\|\bar{c}_i - c_j\|^2}.
    \label{eq:aug_cost}
\end{equation}
The second term penalizes pairings whose conditions differ, steering
the solver toward condition-consistent couplings.
 
\paragraph{Energy-weighted target marginal and source condition sampling.}
To obtain an approximate COT coupling between the noise and 
target $\tilde{\pi}$, we re-weight the target measure using the energy function $\mathcal{E}$.
For each target sample we define unnormalized weights
$w_j = e^{-\lambda\,\mathcal{E}(a_j, c_j)}$,
which we use to transform the target probability mass function used in the COT problem to
\begin{equation}
\label{eq:weighted_measure}
    \bar{\mathbf{a}}=\displaystyle\sum_{j=0}^m p_j\delta_{(a_j, c_j)}, \quad p_j=\frac{w_j}{\sum_{k=1}^{m} w_k},
\end{equation}
placing more mass on low-energy
(high-reward) samples.
To approximate the COT plan, the source and target measures should share the same condition marginals \citep{kerrigan2024dynamic}.
Since the weighted target has condition marginal
$\sum_j p_j\,\delta_{c_j}$, the source
conditions $\bar{c}_i$ used in \eqref{eq:concat} must be drawn
from this same tilted distribution. We therefore sample each source
condition independently as $\bar{c}_i \sim \sum_{j=1}^mp_j\delta_{c_j}$.
Combining the source conditions with \eqref{eq:concat} and \eqref{eq:weighted_measure} we get the following measures $\bar{\mathbf{z}} = \sum_{i=0}^m\frac{1}{m}\delta_{\bar{z}_i}$ and $\bar{\mathbf{a}} = \sum_{j=0}^m p_j \delta_{\bar{a}_j}$.

Putting everything together, the coupling used for training is obtained
by solving the entropic OT problem $T_\varepsilon(\bar{\mathbf{z}}, \bar{\mathbf{a}})$ at each mini-batch. Eventually, at each training step we sample noise-action pairs from the optimal coupling $\Gamma^*$ that minimizes $T_\varepsilon(\bar{\mathbf{z}}, \bar{\mathbf{a}})$ and use \eqref{eq:cfm_loss} to train a flow. We term this loss \textit{wCOT-CFM}. In \Cref{fig:moons} we can see that flows trained using wCOT-CFM successfully sample high energy samples with near straight lines and good low-NFE accuracy compared to an energy weighted flow matching approach \citep{zhang2025energyweighted}.

\subsection{Policy post-training using wCOT-CFM}
\label{sec:otql_rl}

We next apply wCOT-CFM to the problem of RL post-training. We assume access to a pre-trained flow policy $v_{\bar{\theta}}$ that samples from a distribution $\pi_{\bar{\theta}}$ and we train it using wCOT-CFM to extract a new policy that generates samples from 
$\pi_{\theta}(a\mid s)\propto\pi_{\bar{\theta}}(a\mid s)e^{\lambda A(a, s)}$.

\paragraph{Training the critic.} 
We assume access to a buffer $\mathcal{B}$ that may contain transitions from an expert or semi-expert offline dataset $\mathcal{D}_{\rm off}$ that was used to train the base flow $v_{\bar{\theta}}$ and policy rollouts using $v_\theta$. We train a critic $Q_\phi(s, a)$ using a Temporal Difference (TD) loss:
\begin{equation}
\label{eq:td_loss}
    L_Q(\phi) = \mathbb{E}_{(s_t, a_t, r_t, s_{t+1})\sim \mathcal{B},\; a_{t+1} \sim \pi_\theta(\cdot\mid s_{t+1})}\big[ \big( Q_\phi(s_t, a_t) - (r_t + \gamma Q_{\bar{\phi}}(s_{t+1}, a_{t+1}))\big)^2\big]
\end{equation}
where $a_{t+1} \sim \pi_\theta(\cdot\mid s_{t+1})$ implies $a_{t+1}\leftarrow \operatorname{ODE}(v_\theta, s_{t+1}, z)$ which is the \texttt{euler} integration of the flow $v_\theta$ with condition $s_{t+1}$ and source noise sample $z\sim \mathcal{N}(0,I_d)$. The target $y = r + \gamma Q_{\bar{\phi}}(s_{t+1}, a_{t+1})$ is the Bellman target commonly used in RL \citep{park2025flow, tiofack2025guided, pmlr-v305-wagenmaker25a} and $Q_{\bar{\phi}}$ is a target network whose parameters are updated using an exponential moving average filter. 

\begin{wrapfigure}{R}{.6\textwidth}
\begin{minipage}{\linewidth}
    \vspace{-30pt}
\begin{algorithm}[H]
\caption{Optimal Transport Q-Learning (OTQL)}
\label{alg:otql}
\begin{algorithmic}
\State \textbf{Input:} Base policy $v_{\bar{\theta}}$, Buffer $\mathcal{B}$, Energy scale $\lambda > 0$, Target EMA $\tau \in (0,1)$
\State \textbf{Initialize:} $v_\theta \gets v_{\bar{\theta}}$, \quad $Q_\phi$, \quad $Q_{\bar{\phi}} \gets Q_\phi$
\State Populate $\mathcal{B}$ via initial environment rollouts using $v_{\bar{\theta}}$

\For{each training iteration}
    \State Sample mini-batch $\{(s_j, a_j, r_j, s_j')\}_{j=1}^m \sim \mathcal{B}$

    \LComment{\color{dimblue} Critic $Q_\phi$ update}
    \State $z\sim\mathcal{N}(0, I_d), \quad a_j' \leftarrow \operatorname{ODE}(v_\theta, s_j', z)$
    \State Update $\phi$: $\sum_{j=1}^m \big( Q_\phi(s_j, a_j) - r_j + \gamma Q_{\bar{\phi}}(s'_j, a'_j) \big)^2$
    \State $\bar{\phi} \gets \tau\phi + (1-\tau)\bar{\phi}$

    \LComment{\color{dimblue} Flow $v_\theta$ update using wCOT-CFM and $A(s,a)$}
    \State $z_k\sim\mathcal{N}(0, I_d), \quad \hat{a}_j^k \leftarrow \operatorname{ODE}(v_\theta, s_j', z_k)$
    \State $A_j(s_j, a_j) \gets Q_\phi(s_j, a_j) - \sum_{k}Q_\phi(s_j, \hat{a}_j^k)$
    \State $p_j \gets \frac{e^{\lambda A_j}}{\sum_k e^{\lambda A_k}}$
    \State $\bar{\mathbf{a}} \gets \sum_{j=1}^m p_j \delta_{\bar{a}_j}, \quad \text{with} \;\; \bar{a}_j = [a_j \parallel \alpha s_j]$
    \State Sample $z_i \sim \mathcal{N}(0, I_d)$ and $\bar{c}_i \sim \sum_{j=1}^mp_j\delta_{c_j}$
    \State $\bar{\mathbf{z}} \gets \frac{1}{m}\sum_{i=1}^m \delta_{\bar{z}_i}, \quad \text{with} \;\; \bar{z}_i = [z_i \parallel \alpha \bar{c}_i]$
    \State Sample $(z, a) \sim \Gamma^*$,    where $\Gamma^* \gets \argmin T_\varepsilon(\bar{\mathbf{z}}, \bar{\mathbf{a}})$
    \State Update $\theta$: $\mathcal{L}_{\text{CFM}}(\theta)$ 

    \LComment{\color{dimblue} Online Data Collection}
    \If{online environment is available}
        \State Rollout using  $v_\theta$ and add new transitions to the buffer $(s, a, r, s')\to \mathcal{B}$
    \EndIf
\EndFor
\end{algorithmic}
\end{algorithm}
\vspace{-20pt}
    \end{minipage}
\end{wrapfigure}

The actions $a_t$ are typically single-step actions; however they can also be thought of as action chunks $a_t = (a^t_0, a^t_1, \dots, a^t_{H-1})$ of horizon $H$ where $a^t_i$ is the action at $t=t+i$ generated as part of an action chunk at time $t$ \citep{li2026reinforcement}. Then we can simply substitute the instantaneous reward $r_t$ in \eqref{eq:td_loss} with the discounted sum of rewards over the chunk $\sum_{t'=t}^{t+H-1} \gamma^{t'-t}r_{t'}$ making the bellman target an unbiased value function approximator over action chunks. This is important as most pretrained diffusion and flow policies usually operate on chunked actions \citep{chi2023diffusion, intelligence2025pi06vlalearnsexperience, bjorck2025gr00t, shukor2025smolvla, kim2026cosmos} and we therefore adopt the chunked Q function and TD loss throughout this work. 

\paragraph{Training the flow policy.} 
Given the base policy $v_{\bar{\theta}}$ we first initialize $v_\theta$ with $\theta=\bar{\theta}$. Before starting training of the policy and the critic we rollout the base policy in the environment for a small number of episodes and populate the buffer $\mathcal{B}$ (or augment it if it contains $\mathcal{D}_{\rm off}$) with the transitions collected.  We then sample a batch of state-action-reward transitions and compute the advantage function $A(s_t, a_t)= Q(s_t, a_t) - \hat{V}(s_t)$, where $\hat{V}(s_t)$ is a value function approximation calculated as $\hat{V}(s_t)=\mathbb{E}_{a\sim \pi_\theta(\cdot\mid s_t)}\big[Q(s_t, a)\big]$ \citep{nair2020awac}. Lastly, we use the advantage values as a negative energy and calculate the wCOT coupling which is used in $\mathcal{L}_\text{CFM}$ to update $\theta$. The full training pipeline can be seen in \Cref{alg:otql}.

\paragraph{Practical implementations.}
wCOT-CFM trains an energy-informed flow by modifying the coupling used in \eqref{eq:cfm_loss}. Sampling noise-action pairs from the optimal coupling $\Gamma^*$ will possibly lead to some actions in the batch to be dropped and others to be duplicated.
When the $Q$ function has sharp peaks, sampling from the optimal coupling may significantly reduce the effective batch size and increase the loss variance. 
Therefore, we make two practical modifications to overcome this challenge: 1) we clamp the weights $w_j = \min(e^{\lambda\mathcal{E}}, w_{max})$ in \eqref{eq:weighted_measure} and 2) at inference time and for the Bellman target computation, we generate $N$ candidate actions from the flow and select the one with the highest $Q$-value \citep{chen2023offline, li2026reinforcement}. 
This is equivalent to approximately sampling from a sharpened version of the policy $\pi_\theta$, and can be seen as a complementary mechanism to the OT-based steering: while wCOT-CFM shapes the flow's trajectory distribution toward high-advantage regions, rejection sampling provides a lightweight refinement that further concentrates samples near the mode of the advantage-weighted distribution. 
We find that N=5 candidates suffice to obtain meaningful gains, adding only negligible computational overhead relative to the ODE integration cost. 
Although we employ weight clamping for all tasks, we only use rejection sampling in scenarios with relatively large chunk sizes and peaked $Q$ functions. 
See \Cref{sec:appendix_ablation} for an ablation of the most important OTQL hyper-parameters.


\section{Experiments}
\vspace{-6pt}
\label{sec:experiments}
We evaluate the ability of OTQL to train flow policies in offline, online, and offline-online RL settings across a diverse set of 42 simulation tasks and 4 real-world tasks. For the simulations, we utilize the OGBench benchmark \citep{ogbench_park2025} for offline and offline-online experiments, alongside two MuJoCo tasks \citep{todorov2012mujoco} from the LeRobot codebase \citep{cadene2024lerobot} for online post-training. We compare OTQL against several baselines: DSRL \citep{pmlr-v305-wagenmaker25a}, which has proven highly effective in robotics; Q-Chunking (QC) \citep{li2026reinforcement}, which shares OTQL's chunking and rejection sampling mechanisms; and the fast flow-based methods FQL \citep{park2025flow} and QAM-F \citep{li2026q}. We also evaluate Energy Weighted Flow Matching (EWFM) \citep{zhang2025energyweighted}, a method that, like OTQL, connects to the importance sampling framework. Some offline RL baselines are diffusion-based (QSM \citep{pmlr-v235-qsm}) or flow adaptations of diffusion-based methods (IFQL \citep{hansen2023idql}). FAWAC, CGQL, and FEdit are baselines considered and detailed in \citep{li2026q}. 

\subsection{Offline RL in simulation}
\label{sec:experiments_offline_rl}

\begin{table}[t]
    \centering
    \resizebox{\linewidth}{!}{
    \begin{tabular}{l ccccccc cc c}
        \toprule
        & \multicolumn{7}{c}{NFE $\geq$ 10} & \multicolumn{2}{c}{NFE = 1} & \multicolumn{1}{c}{NFE = 2} \\
        \cmidrule(lr){2-8} \cmidrule(lr){9-10} \cmidrule(lr){11-11}
        Environment & FAWAC & DSRL & QSM & CGQL & FEdit & QAM & IFQL & FQL & QAM-F & \textcolor{dimblue}{OTQL} \\
        \midrule
        \texttt{antmaze-large}      & 0.17 & 0.61 & \textbf{0.9} & 0.76 & 0.58 & 0.81 & 0.36 & 0.76 & 0.83 & \underline{0.88} \\
        \texttt{antmaze-giant}       & 0 & 0.03 & \textbf{0.24} & 0 & 0.02 & 0.18 & 0.01 & 0 & 0.12 & \underline{0.19} \\
        \texttt{humanoidmaze-medium}  & 0.24 & 0.53 & \underline{0.82} & 0.6 & 0.22 & 0.67 &\textbf{0.86} & 0.68 & 0.65 & \underline{0.82} \\
        \texttt{humanoidmaze-large}   & 0 & 0.03 & 0.06 & 0.05 & 0.03 & 0.11 & \textbf{0.24} & 0.09 & 0.12 & \underline{0.19} \\
        \texttt{cube-double}          & 0.02 & \textbf{0.74} & 0.33 & 0.38 & 0.4 & 0.64 & 0.11 & 0.46 & 0.65 & \underline{0.68} \\
        \texttt{cube-triple}         & 0 & 0.01 & \underline{0.06} & \textbf{0.08} & 0.02 & 0.03 & 0 & 0.03 & 0.03 & 0.04 \\
        \texttt{scene}               & 0.38 & \textbf{0.99} & 0.78 & 0.38 & 0.62 & 0.97 & 0.84 & 0.78 & 0.95 & \textbf{0.99} \\
        \texttt{puzzle-3x3}          & 0.03 & 0.87 & 0.57 & 0.48 & \textbf{0.99 }& \textbf{1} & \textbf{1} & 0.7 &\textbf{ 0.99} & \textbf{0.99} \\
        \midrule
        \textbf{Average}                         & 0.11 & 0.48 & 0.47 & 0.34 & 0.36 & \underline{0.55} & 0.43 & 0.44 & 0.54 & \textbf{0.59} \\
        \bottomrule
    \end{tabular}
    }
    \vskip 0.1in
    \caption{Offline RL results on OGBench. Each environment has 5 tasks associated with it.}
    \label{tab:offline_results}
    \vspace{-20pt}
\end{table}

We first test the ability of OTQL to train flow policies in offline settings, on the OGBench benchmark \citep{ogbench_park2025}. For all tasks, a dataset of suboptimal and mixed-quality demonstrations is provided along with sparse rewards (0 if an action solves the tasks, -1 otherwise) and the goal is to extract meaningful behaviors for solving the task from it. The main results across 40 tasks and 9 baselines are reported in \Cref{tab:offline_results}, where the methods are grouped based on the NFE required to sample one action from the flow or diffusion policy. We follow the same experiment setup as in \citep{li2026q} and utilize action chunking for the tasks \texttt{cube-double, cube-triple, scene} and \texttt{puzzle}. For OTQL we also employ rejection sampling only in the aforementioned tasks with $N=5$.

From \Cref{tab:offline_results} we observe that OTQL performs equally well and in some tasks even outperforms state-of-the-art methods that require multiple NFE for inference. Overall, OTQL achieves the highest success rate on average and outperforms fast inference methods like FQL and QAM-F by at least $5\%$. This showcases the ability of OTQL to train policies from scratch given suboptimal demonstrations, while maintaining competitive inference speed. Offline pretraining can be very effective when training policies from data sources that contain demonstrations of mixed quality \citep{robomimic2021} or when using the robot's failed rollouts \citep{intelligence2025pi06vlalearnsexperience}. We utilize OTQL in an offline setting extensively in the real-robot tasks (\Cref{sec:experiments_real}).

\subsection{Offline-online and online RL in simulation}
\label{sec:experiments_online_rl}

\begin{figure}[t]
    \centering
    \begin{minipage}{0.49\textwidth}
        \centering
        \includegraphics[width=\linewidth]{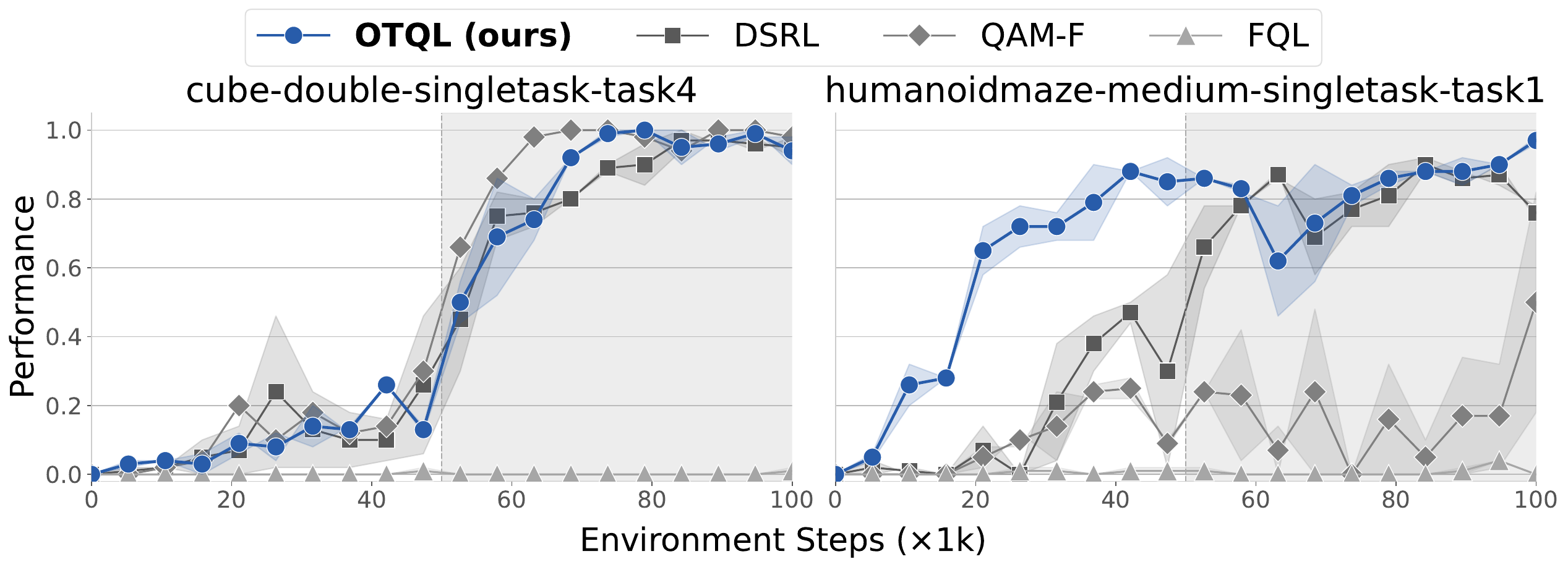}
        \caption{OTQL is effective in offline-online RL settings}
        \label{fig:offline_online}
    \end{minipage}\hfill
    \begin{minipage}{0.49\textwidth}
        \centering
        \includegraphics[width=\linewidth]{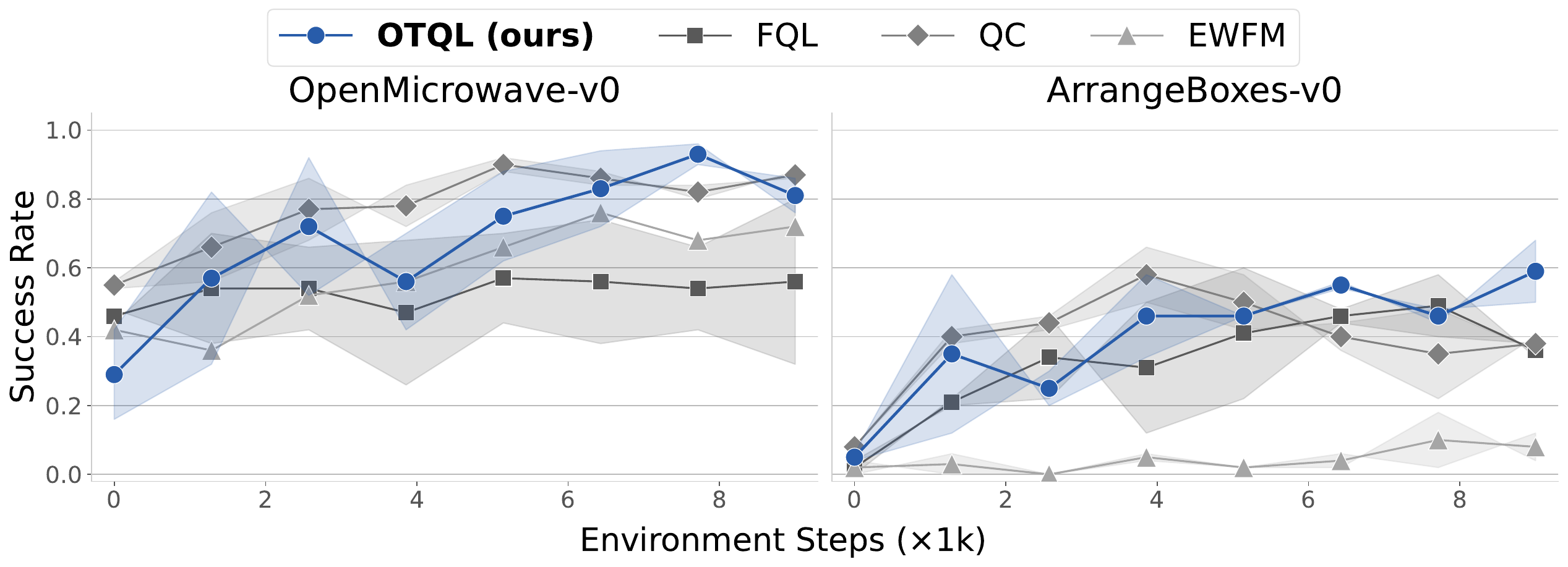}
        \caption{OTQL enables adaptation and acceleration in online RL settings}
        \label{fig:online} 
    \end{minipage}
    \vspace*{-8pt}
\end{figure}

We further explore the ability of OTQL to pre-train a policy using offline RL and then finetune it using online RL. Additionally, we test the online adaption capabilities of OTQL when provided with a pre-trained policy that has been trained on a limited amount of demonstrations. 
\paragraph{Offline-to-online adaptation.} We test offline-to-online adaptation in two OGBench environments, one in which we use action chunking (\texttt{cube-double}) and one in which no action chunking is employed (\texttt{humanoidmaze-medium}). DSRL, FQL and QAM-F are used as baselines as they are among the best performing methods in the offline RL experiments and FQL and QAM-F explicitly accelerate inference. All policies are pre-trained for 500k steps and then finetuned with online RL for 500k steps with an update-to-data ratio (UTD) of 1. The training curves are shown in \Cref{fig:offline_online}. We observe than in the \texttt{cube} task where action chunking is employed all methods have similar performance, with a clear performance gain when switching from offline to online RL. However, in the \texttt{humanoidmaze} task we observe that OTQL is the only method reaching near $90\%$ with offline pre-training, while DSRL is the only baseline matching OTQL's performance during online training. FQL and QAM-F, which have similar inference time to OTQL,  fail to surpass a success rate of $50\%$.
\paragraph{Online finetuning.} We further test if OTQL can steer flow policies pre-trained on a limited amount of demonstrations. More specifically, we evaluate OTQL, QC, FQL and EWFM on two Mujoco tasks, namely \texttt{OpenMicrowave-v0} and \texttt{ArrangeBoxes-v0} (see \Cref{sec:appendix_implementations_details}). We use action chunking for all methods with a chunk size of 10. The pre-trained policy is first rolled-out in the environment for 2000 steps and the transitions are added to the replay buffer $\mathcal{B}$ alongside the pre-training demonstrations $\mathcal{D}_{\rm off}$, similar to \citep{pmlr-v305-wagenmaker25a}. We then train the policies for a total of 8000 steps using $\text{UTD}=10$. For OTQL, we only use rejection sampling on \texttt{ArrangeBoxes-v0} with $N=5$ candidates. Overall, OTQL provides the most effective online adaptation. Compared to high-NFE baselines, OTQL ($\text{NFE}=3$, $N=5$) outperforms EWFM ($\text{NFE}=10$, $N=1$) and performs as well as QC ($\text{NFE}=10$, $N=32$) while requiring only a fraction of the inference time. At the same time, it maintains a significant performance edge over other fast methods, achieving a maximum success rate that is on average $15\%$ higher than FQL. While the randomized box locations in \texttt{ArrangeBoxes-v0} limit the absolute gains of all methods, OTQL consistently delivers the best combination of task success and inference speed.

\subsection{Policy steering and acceleration in real-world tasks}
\label{sec:experiments_real}

\begin{figure}[t]
    \centering
    \includegraphics[width=0.8\linewidth]{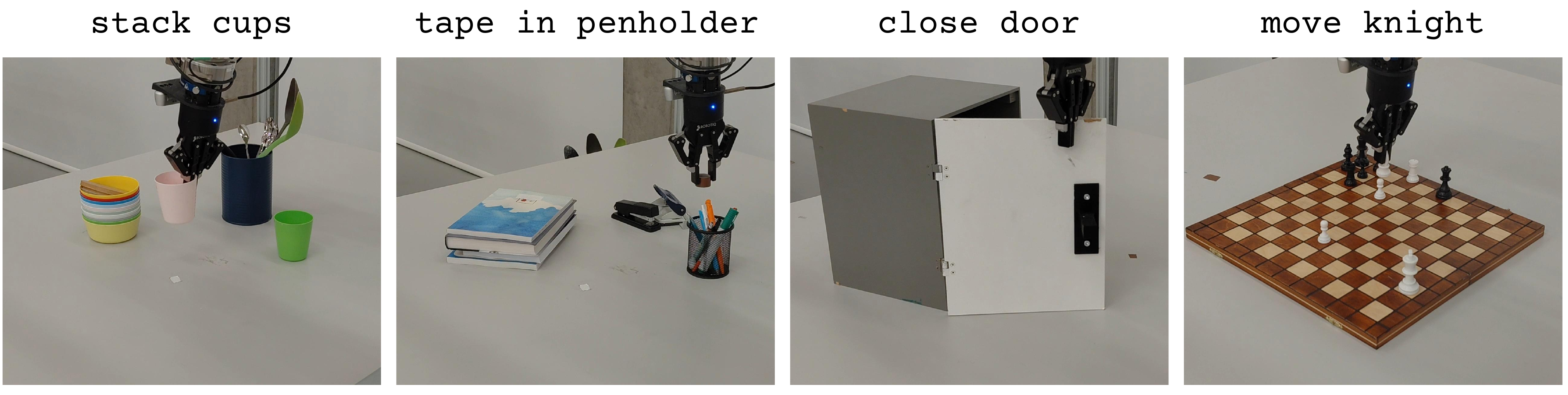}
    \caption{real-wrold tasks used for the evaluation of OTQL}
    \vspace*{-1em}
    \label{fig:tasks}
\end{figure}

We next evaluate OTQL's ability to fine-tune and accelerate both single-task and flow-based VLA policies under challenging real-world conditions. We evaluate our method across a suite of four tasks (\Cref{fig:tasks}), selected to represent varied manipulation paradigms. By spanning stacking, accurate manipulation, contact-rich pushing, and precise pick-and-place maneuvers, we aim to demonstrate OTQL's robustness across diverse skills and precision requirements. All evaluations operate under a maximum episode length of 150 steps with sparse rewards ($r_t = 0$ if the policy solves the task at timestep $t$, and $r_t=-1$ otherwise \citep{intelligence2025pi06vlalearnsexperience})

\paragraph{OTQL for real-world policies.} To test how well OTQL can adapt single-task policies, we train a base flow policy on each of the first three tasks in \Cref{fig:tasks}, using CFM with 10 demonstrations. The base policy typically achieves low performance ($\leq50\%$) as the tasks require precise behaviors to be solved. Instead of doing online RL, we choose to adapt the policies by training in a semi-offline setting every 10 demonstrations. 
\begin{wrapfigure}{r}{0.6\textwidth}
    \vspace*{-1em}
    \centering
    \includegraphics[width=\linewidth]{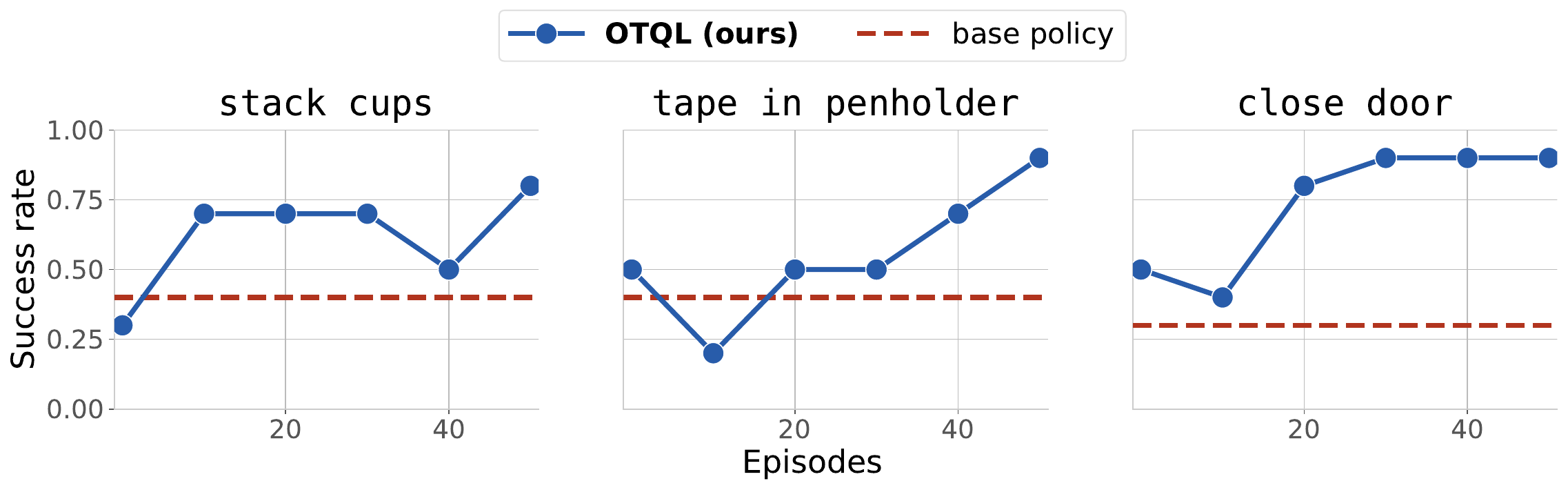}
    \caption{OTQL adaptation of base flow policies on 3 tasks.}
    \vspace*{-1em}
    \label{fig:online_real}
\end{wrapfigure}
We start by collecting 10 rollouts from the base policy and train using offline RL on the expert demonstrations and the collected data for 20k steps. We then iteratively rollout the fine-tuned policy for 10 episodes and train with OTQL for 20k steps until the buffer contains a total of 50 episodes. We found that this approach significantly reduced the amount of time a human needed to supervise the robot, as no training was performed during the rollouts. Also this approach enables the policy to be deployed in higher frequencies, allowing adaptation in tasks that require fast motions and responsiveness. More details on the experiments setup can be found in \Cref{sec:appendix_implementations_details}. Overall, we observe from \Cref{fig:online_real} that OTQL is able to adapt the base policies by increasing average success rates from $36\%$ to $86\%$, while reducing the NFE needed for inference from $10$ to only $3$, offering a significant inference speed-up.

\paragraph{OTQL for VLAs.} Finally, we highlight the efficacy of our approach by fine-tuning a popular, state-of-the-art VLA \citep{shukor2025smolvla} on the \texttt{tape in penholder} and \texttt{move knight} tasks. We first use SFT to fine-tune the SmolVLA action head with 10 demonstrations, as the publicly available pre-trained checkpoint failed zero-shot on both tasks.  
\begin{wraptable}{r}{0.5\textwidth}
    \centering
    \resizebox{\linewidth}{!}{
    \begin{tabular}{lcc}
        \toprule
        \textbf{Task} & \textbf{SmolVLA} & \textbf{SmolVLA-OTQL} \\
        \midrule
        \texttt{tape in penholder} & 14/30 & \textbf{24/30} \\
        \texttt{move knight} & 9/30 & \textbf{22/30} \\
        \bottomrule
    \end{tabular}
    }
    \caption{Comparison of SmolVLA performance on two real-world tasks trained with SFT and OTQL.}
    \label{tab:smolvla}
    \vskip -1em
\end{wraptable}
After training with SFT, the policy adapts to the tasks but performance is still suboptimal with success rates below $50\%$, as can be seen in \Cref{tab:smolvla}. We utilize OTQL to adapt SmolVLA on a set of 30 rollouts collected form the fine-tuned VLA. 
From \Cref{tab:smolvla}, we see that this is enough to adapt it to the two hardest out of the four tasks, increasing the average success rate by an average of $37\%$ while bringing down its inference NFE to 3. We note that similarly to the single-task case, online finetuning could be more effective, however, given the time each training iteration takes, training online would result in intermittent robot motions and could possibly compromise success. Throughout these experiment, we kept the VLM of SmolVLA frozen as fully fine-tuning it in a very low-data regime could risk severe overfitting \citep{driess2026knowledge}. 

\section{Conclusions}
\vspace{-6pt}
\label{sec:conclusions}

We present OTQL, a method that steers and accelerates flows. We employ weighted COT couplings and CFM to fine-tune policies, leading to an efficient and easy to implement training algorithm. We have shown that OTQL outperforms other post-training methods and that it can adapt \textit{and} accelerate single-task flow policies and VLAs in the real-world with a limited interaction budget. OTQL can act as the bridge between large-scale pretraining and efficient real-time performance on new tasks that the pretrained policy solves only partially.  

\paragraph{Limitations.} OTQL is effective in adapting flow policies in both real and simulation environments using a single mechanism for steering and acceleration, however, there are certain limitations that need to be highlighted. First,the policy can't be treated as a black box, contrary to other methods \citep{pmlr-v305-wagenmaker25a, ankile2025imitation}, since OTQL assumes the policy is a neural ODE-based generative model. The complexity of post-training large flow policies can be eliminated with techniques like LoRA \citep{hu2022lora}, however not all VLAs or WAMs are compatible with our method as formulated in this paper. Second, the inference speed-up and the adaptation become more modest as the dimensionality of the problem increases, as minibatch OT is prone to larger errors in higher dimensions. Lastly, our method inherits many of the problems typically encountered in real-world RL methods such as manual episode labeling and manual environment resets after every episode.


\clearpage


\bibliography{references}  

\clearpage
\appendix

\section{Optimal Transport}
\label{sec:appendix_ot}
The static OT problem aims to find a coupling of minimal cost between two distributions. Given a cost function $C(\cdot, \cdot):\mathbb{R}^d \times\mathbb{R}^d\rightarrow\mathbb{R}$, we search for a solution to the optimization problem:
\begin{align}
    \label{eq:static_ot}
    \mathrm{OT}(p_z,p_a) = \inf_{\pi \in \Pi(p_z,p_a)} \mathbb{E}_{(z,a)\sim\pi}[ C(z,a)],
\end{align}
where $\Pi(p_z,p_a)$ denotes the space of probability measures whose left and right marginals equal $p_z$ and $p_a$, respectively. Intuitively, we seek a way to (stochastically) transport particles distributed according to $p_z$ so as to make them distributed according to $p_a$, minimizing the total cost of transportation. The infimum in \eqref{eq:static_ot} is called the OT cost, and a minimizer is called an OT plan (under some conditions, \eg, squared Euclidean cost and $p_z$ absolutely continuous with finite variance), the infinimum is achieved and the minimizer is unique \cite{brenier1991polar}). In the case where $C(z,a)=||z - a||^2$ is the squared Euclidean distance, the square root of the distance in \eqref{eq:static_ot} is called the 2-Wasserstein distance and denoted $W_2(p_z,p_a)$. 

The 2-Wasserstein distance can also take a dynamic form, called the Benamou-Brenier form~\citep{benamou2000computational}, which involves a flow vector field $u(t, x)$, which transforms one probability distribution to the other:
\begin{align}
    \label{eq:dynamic_ot}
    W_2(p_z, p_a) = \inf_{p(t, x), u(t,x)} \int_{\mathbb{R}^d} \int_0^1 p(t, x) \|u(t, x)\|^2 \, dt \, dx,
\end{align}, 
where $p(t, x)$ is the probability path generated by $u(t, x)$ and has marginal distributions $p(0,\cdot) = p_z,\ p(1, \cdot) = p_a$. The minimizer of \eqref{eq:dynamic_ot} is a vector field that produces sample paths of minimum length, which subsequently forces them to be straight and thus easier to simulate
with limited \texttt{euler} steps.

In the case of conditional distributions with conditions $c\in \mathcal{C}$ a relaxed version of the static OT problem is the conditional Kantorovich problem~\citep{hosseini2023conditional}:
\begin{align}
    \label{eq:static_cot}
    \inf_{T \in \Pi_c} \int_{\mathbb{R}^d \times \mathbb{R}^d} c((z, \tilde{c}), (a, c)) \, dT((z, \tilde{c}), (a, c)),
\end{align}
where $\tilde{c}$ are source conditions and $\Pi_c:=\{T\in\Pi\ |\ ((z, \tilde{c}), (a, c))\sim T, \text{such that }\tilde{c}=c\}$ is the space of conditional distributions where the conditions are equal. \citet{kerrigan2024dynamic} introduced an equivalent Benamou-Brenier dynamic form for the conditional OT problem.

\section{Related Work}
\label{sec:appendix_related_work}

\paragraph{Diffusion/flow models for robot control.}
Diffusion \citep{ho2020denoising, song2020score} and flow \citep{lipman2022flow} models have demonstrated excellent multimodal and high-dimensional distribution learning capabilities \citep{ho2022video, agarwal2025cosmos, rombach2021highresolution, saharia2022photorealistic}, which rendered them a state-of-the-art model choice for action generation in robotics \citep{urain2024deep}. \citet{chi2023diffusion} first introduced diffusion models for robot control via behavior cloning, by learning the conditional distribution of action chunks instead of single-step actions. Other early works have investigated the use of diffusion models as planners \citep{pmlr-v162-janner22a} that use rewards to steer the generation of actions using diffusion guidance \citep{ho2022classifier}. Diffusion and flow models have also proven effective policy representations in scenarios with multimodal observations, such as forces \citep{hou2025adaptive, xue2025reactive} and point-clouds \citep{chisari2024learning, ze20243d}.     

Diffusion and flow models, owing to their ability to sample from complex distributions with flexible conditioning, are very popular components for action generation in more complicated architectures used for generalist policies. More specifically,  most VLA architectures \citep{shukor2025smolvla, BlackK-RSS-25, pmlr-v305-black25a, bjorck2025gr00t, reuss2025flower}, either export features from the VLM or generate next tokens autoregressively with the VLM which are used for conditioning a flow- or diffusion-based action head. The same principle has also been extended to WAMs \citep{kim2026cosmos, pai2025mimic}, which use a video or world model backbone instead of a VLM. Other architectures, train diffusion models which generate both future images and actions \citep{LiS-RSS-25, zhu2025unified} with the aim of improving scalability through large-scale video pretraining.   

\paragraph{Inference acceleration of generative policies.}
Diffusion and flow model inference requires integration of an ODE or SDE and therefore multiple neural function evaluations for generating one sample from the target distribution. As a consequence, accelerating inference by reducing the amount of integration steps needed for the SDE/ODE integration is relevant across many fields \citep{zhang2025turbodiffusion, Cheng_2025_ICCV, geng2026mean}. Especially in robotics, high inference speeds are crucial for ensuring reactive and continuous robot motions \citep{black2026real}. Some methods have achieved one-step inference using consistency objectives \citep{prasad2024consistency, wang2024one, lu2024manicm} at the cost of more expensive training. Other methods avoid computationally complex training by using the flow variance to adapt the integration steps \citep{hu2024adaflow} or by using COT couplings to straighten the integration paths \citep{sochopoulos2025fast}. More efficient training, however, comes at the cost of one step inference quality, as they typically require 2-3 NFE in action generation problems. Instead of modifying the training objective of the generative policies, \citet{black2026real} propose a practical post-processing alternative to overcoming intermittent robot motions, by generating the next action chunk while the current one is already being executed by the robot. 

\paragraph{RL adaptation of flow policies.}
The same qualities that led to the success of diffusion and flow models as robot policies have proven useful in reinforcement learning too \citep{pmlr-v235-qsm, li2024learning, hansen2023idql}. Flow models have unlocked new opportunities for policy and value learning, particularly in offline RL settings, where the policy and critic have access only to a static dataset of rollouts collected by a base policy \citep{fujimoto2021a}. The goal in offline RL is to sample from the tilted distribution that is the closed form solution of the KL regularized problem \citep{peters2007reinforcement, peters2010relative}. Sampling from this distribution has been achieved using advantage and critic weighted flow matching objectives \citep{zhang2025energyweighted, tiofack2025guided}, adjoint matching \citep{li2026q}, rejection sampling \citep{hansen2023idql, li2026reinforcement}, classifier guidance \citep{huang2025offlinetoonline} and by using the relative trajectory balance objective for diffusion models \citep{venkatraman2024amortizing}. Since all these methods require multi-step integration for action generation, \citep{park2025flow} distills the behavior of a flow policy to a single-step guassian policy while also optimizing a $Q$ loss.  

In robotics, access to a suboptimal pre-trained base policy is assumed and the goal is to extract a policy that samples high value actions and actions that also have a high density under the base policy density. \citet{pmlr-v305-wagenmaker25a} learn a policy that generates noise samples for the base policy, using RL. Other methods focus on guidance \citep{du2026dynaguide} or residual policies correcting the actions generated by the base policy \citep{ankile2025residual, ankile2025imitation}.

\section{Ablation}
\label{sec:appendix_ablation}

\begin{figure}[h]
    \vspace*{-1em}
    \centering
    \includegraphics[width=\linewidth]{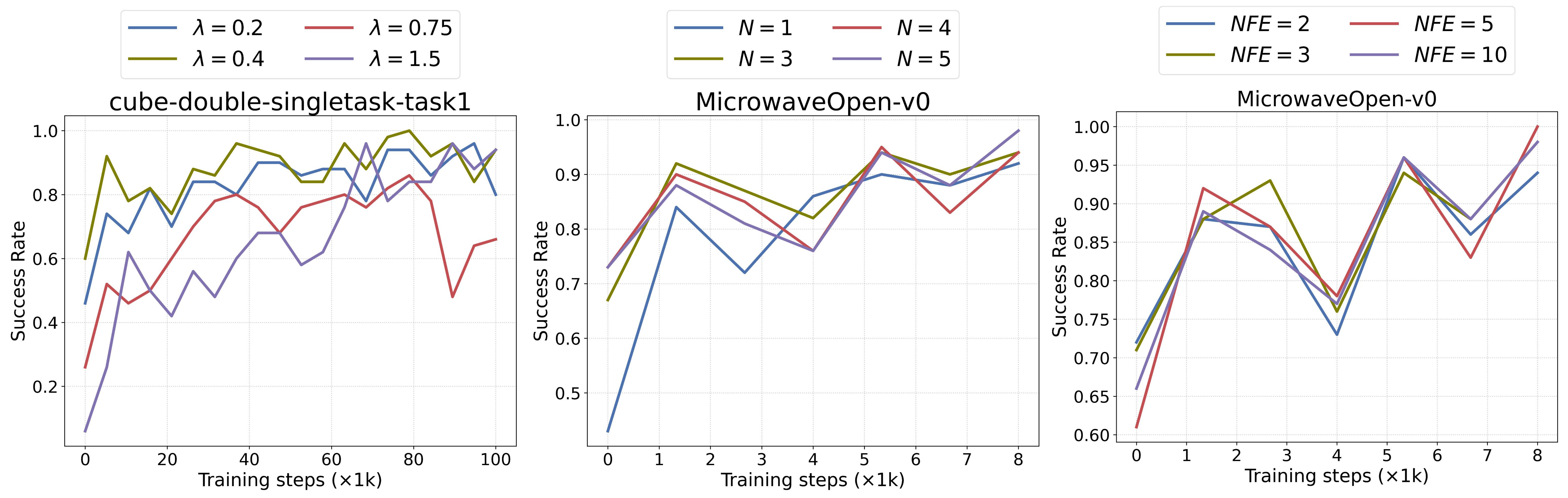}
    \caption{Ablation of OTQL for $\lambda, N$ and inference steps NFE}
    \label{fig:ablation}
\end{figure}

In this section we ablate the most important hyper-parameters of OTQL and provide some insight on how to tune them. In \Cref{fig:ablation} we show the effects of the temperature parameter $\lambda$, the number of action candidates used for rejection sampling $N$ and the NFE used for sampling actions.  

\paragraph{Temperature $\lambda$.} OTQL uses importance weights to assign more mass to low energy samples in the batch while calculating the empirical COT coupling. If the temperature $\lambda$ is too high then there will be only a few samples in the batch that will be assigned a significant weight leading to a transport plan $\Gamma^*$ that is concentrated around these few points. This means that when we sample from $\Gamma^*$ most of the batch will be potentially discarded, leading to a significantly reduced effective batch size. This can make the training slower to converge as depicted in \Cref{fig:ablation} for $\lambda=1.5$. The policy still ends up performing adequately, however many more training iterations are required until convergence. 

On the other side, if $\lambda$ is too small then potentially all samples in the batch could be assigned similar weights. This can lead to a COT coupling that does not prioritize high advantage samples and therefore fails to achieve high success rates. In practice, we choose $\lambda$ to be between $[0.4, 1.0]$ in tasks where the chunk size is 10 and we have approximately 100-200 maximum episode steps. Other tasks and experiment choices can greatly impact the optimal value for $\lambda$. 

\paragraph{Rejection sampling candidates $N$.} As explained in \Cref{sec:otql_rl} in certain tasks we employ rejection sampling with a relatively low number of candidates ($N=5$). This decision is grounded on the fact that we have to choose a moderate $\lambda$ value to avoid the problem described before, which might lead to difficulties in adapting policies with critics that have sharp peaks, or when high advantage actions are sparse. In \Cref{fig:ablation} we show how rejection sampling assists our method. We use a moderate $\lambda$ value which gives us a good approximation of the tilted distribution $\tilde{\pi}$ and then use rejection sampling to make it more concentrated around high value actions. Eventually, rejection sampling assists in earlier training iterations resulting in more successful rollouts in the buffer which in turn assists OTQL. We see that towards the middle and end of training rejection sampling does not further improve the performance compared to vanilla OTQL. This shows that OTQL does indeed take care of most of the adaptation and rejection sampling acts only as a lightweight filtering. This is also the reason why 3-5 candidates are enough compared to QC which uses 32 in many tasks on OGBench \citep{li2026reinforcement}.  

\paragraph{NFE for flow inference.} OTQL uses COT coupling both for straightening the integration paths of the flow and for distribution steering. Apart from the competitive performance shown in \Cref{sec:experiments} with 2 or 3 NFE during inference and training, we also show that flows trained with OTQL perform the same no matter the amount of steps used for inference and training. This can be seen in \Cref{fig:ablation}, where we see that the fine-tuned policy has the same performance overall throughout training for all NFE between 2 and 10. Many flow and diffusion RL methods \citep{pmlr-v235-qsm, ICLR2025_dac, li2026q} require at least 10 NFE for achieving competitive performance, while our method is competitive with $\text{NFE}=2$ and without using distillation \citep{park2025flow, tiofack2025guided}. 

\section{Implementation details}
\label{sec:appendix_implementations_details}

\subsection{Offline RL}
\label{sec:appendix_implementations_details-offline}

In all offline experiments involving OGBench we use the exact same evaluation setup as in \citep{li2026q}. We therefore also employ an ensemble of $K=10$ critic functions $Q^k,\; k=1,\dots,10$ and the pessimistic target value backup \citep{ICLR2025_dac}:
\begin{equation}
    \label{eq:pessimistic_td}
        L_{Q_k} = \mathbb{E}\big[ (Q_\phi^k(s, a) - y)^2\big],
\end{equation}
where $y=r + \frac{\gamma}{K}\big(\sum_kQ^k_{\tilde{\phi}}(s, a) - \rho\sqrt{\sum_k(Q^k_{\tilde{\phi}}(s', a') - \frac{1}{K}\sum_kQ^k_{\tilde{\phi}}(s', a'))^2} \big)$ and $\rho$ is the pessimistic coefficient.

All models used in the policy and the critic were MLPs with parameters shown in \Cref{tab:hyperparameters_offline}.
\begin{table}[htbp]
\centering
\resizebox{0.7\linewidth}{!}{
\begin{tabular}{cc}
\toprule
\textbf{Parameter} & \textbf{Value} \\
\midrule
Batch size & 256 \\
Discount factor ($\gamma$) & \begin{tabular}{@{}c@{}}0.99 for \texttt{puzzle,scene,cube,antmaze-large} \\ 0.995 for \texttt{humanoidmaze,antmaze-giant}\end{tabular} \\
Actor learning rate & $3 \times 10^{-4}$ \\
Critic learning rate & $3 \times 10^{-4}$ \\
Target network update rate ($\lambda$) & $5 \times 10^{-3}$ \\
Critic ensemble size ($K$) & 10 \\
Critic target pessimistic coefficient ($\rho$) & \begin{tabular}{@{}c@{}}0.5 for \texttt{puzzle,scene,cube,antmaze} \\ 0 for \texttt{humanoidmaze}\end{tabular} \\
UTD ratio & 1 \\
Number of offline training steps & $10^6$ \\
Hidden dimensions& 512 \\
Network depth & 4 hidden layers \\
\bottomrule
\end{tabular}
}
\vskip 0.1in
\caption{Parameters used in the offline RL experiments}
\label{tab:hyperparameters_offline}
\vspace{-20pt}
\end{table}

\begin{table}[h]
    \centering
    \resizebox{0.5\linewidth}{!}{
    \begin{tabular}{l c c c c}
        \toprule
        & \multicolumn{4}{c}{OTQL} \\
        \cmidrule(lr){2-5} 
        Environment & $\lambda$  & $w_{max}$ & $N$ & NFE\\
        \midrule
        \texttt{antmaze-large-singletask}        & 2& 3& 1 & 2\\
        \texttt{antmaze-giant-singletask}       & 2& 3& 1& 2\\
        \texttt{humanoidmaze-medium-singletask}  & 5& 1& 1& 2\\
        \texttt{humanoidmaze-large-singletask}  & 5& 1& 1& 2\\
        \texttt{cube-double-singletask}          & 0.4& 10& 5& 2\\
        \texttt{cube-triple-singletask}          & 0.4& 10& 5& 2\\
        \texttt{scene-singletask}               & 0.4& 10& 5& 2\\
        \texttt{puzzle-3x3-singletask}          & 0.4& 10& 5& 2\\
        \bottomrule
    \end{tabular}
    }
    \vskip 0.1in
    \caption{Hyperparameters used for OTQL in the offline RL tasks.}
    \vspace{-20pt}
\end{table}

\subsection{Online RL in simulation}
\label{sec:appendix_implementations_details-online}

\begin{figure}[htbp]
    \centering
    
    \begin{minipage}[c]{0.45\textwidth}
        \centering
        \includegraphics[width=\linewidth]{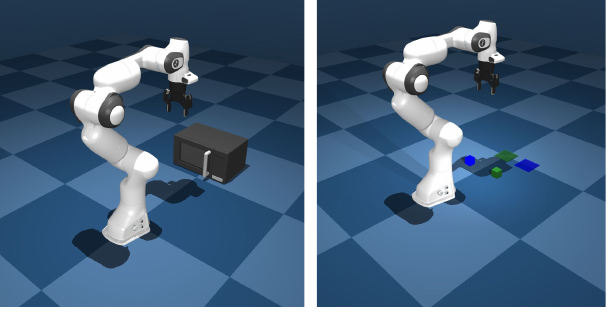}
        \captionof{figure}{MuJoCo tasks for online RL post-training}
        \label{fig:mujoco_tasks}
    \end{minipage}\hfill
    \begin{minipage}[c]{0.50\textwidth}
        \centering
        \resizebox{\linewidth}{!}{%
        \begin{tabular}{cc}
        \toprule
        \textbf{Parameter} & \textbf{Value} \\
        \midrule
        Batch size & 256 \\
        Discount factor ($\gamma$) & 0.99 \\
        Actor learning rate & $3 \times 10^{-4}$ \\
        Critic learning rate & $3 \times 10^{-4}$ \\
        Target network update rate ($\lambda$) & $5 \times 10^{-3}$ \\
        Critic ensemble size ($K$) & 2 \\
        Critic target pessimistic coefficient ($\rho$) & 0.0 \\
        UTD ratio & 10 \\
        Number of online training steps & $8000$ \\
        Number of online training steps before training & $2000$ \\
        Hidden dimensions& 256 \\
        Network depth & 4 hidden layers \\
        \bottomrule
        \end{tabular}%
        }
        \captionof{table}{Parameters used in the online RL experiments}
        \label{tab:online_rl_parameters} 
    \end{minipage}
    
\end{figure}

We ran online RL post-training for OTQL and all the baselines on two MuJoCo tasks visible in \Cref{fig:mujoco_tasks}. The first (\texttt{OpenMicrowave-v0}) requires the robot to open the microwave door and the second (\texttt{ArrangeBoxes-v0}) requires the robot to arrange the two boxes on the correct colored planes. For both tasks we collect demonstrations using a gamepad by controlling only the position of the end-effector and the gripper. We again utilized an ensemble for the critic but not the pessimistic Bellman update. All networks used for the critic and the flow policies where MLPs.

We followed the pipeline presented in \Cref{sec:otql} and first collected transitions of the base policy interacting with the environment for 2k steps. We also included the offline demonstrations collected $\mathcal{D}_{off}$ in the buffer $\mathcal{B}$ and during training we sampled half of the batch from $\mathcal{D}_{off}$ and half from $\mathcal{B}$. We used again an ensemble of critics on all of the environments. 

\begin{table}[h]
    \centering
    \resizebox{\linewidth}{!}{
    \begin{tabular}{@{}l|cc|cc|ccc|ccccc}
        \toprule
        Environment & \multicolumn{2}{c}{FQL} & \multicolumn{2}{c}{QC} & \multicolumn{3}{c}{EWFM} & \multicolumn{4}{c}{OTQL} \\
        \cmidrule(lr){2-3} \cmidrule(lr){4-5} \cmidrule(lr){6-8} \cmidrule(lr){9-12}
         & $\alpha$  & NFE & $N$ &  NFE  & $\lambda$ & $w_{max}$ &  NFE & $\lambda$ & $w_{max}$ & $N$ & NFE  \\
        \midrule
        \texttt{ArrangeBoxes-v0} & 300 & 1 & 32 & 10 & 1.0 & 20 &10& 1.5 &10 & 5 & 3 \\
        \texttt{MicrowaveOpen-v0} & 300 & 1 & 32 & 10 & 1.0 & 20 &10& 1.5 &10 & 1 & 3 \\
        \bottomrule
    \end{tabular}
    }
    \caption{Hyperparameters used for OTQL and the baselines in the online RL tasks.}
    \label{tab:online_training_hyperparameters}
\end{table}

\paragraph{Baselines.} The baselines considered in these experiments are QC \citep{li2026reinforcement}, FQL \citep{park2025flow} and EWFM, which we adapted from \citep{zhang2025energyweighted}. More specifically we used advantage weighting for the weighted flow matching objective instead of critic weighting and employed the same maximum weight clipping technique, like in OTQL. The details of all hyperparameters used in all the baselines can be found in \Cref{tab:online_training_hyperparameters} and the parameters used for training in \Cref{tab:online_rl_parameters}.

\subsection{RL in real-world}
\label{sec:appendix_implementations_details-real}

Similar to the offline and online simulation experiments we use an ensemble of critics for all real-world experiments, but without the pessimistic target update. the experimental protocols for both single-task policies and SmolVLA adaptation are detailed in \Cref{sec:experiments_real}. For single task policies we use 10 expert demonstrations per task and 50 total rollouts (60 episodes) and for the SmolVLA experiments we use 10 expert demonstrations and 30 rollouts (40 episodes). The details of all hyperparameters used in the single-task experiments can be found in \Cref{tab:real_world_rl_parameters} and \Cref{tab:real_world_training_hyperparameters} and for the SmolVLA experiments in \Cref{tab:real_world_rl_parameters_smolvla} and  \Cref{tab:real_world_training_hyperparameters_smolvla}. For all experiments we used LeRobot \citep{cadene2024lerobot}.

\begin{table}[htbp]
    \centering
    
    \begin{minipage}[t]{0.5\textwidth}
        \centering
        \resizebox{\linewidth}{!}{%
        \begin{tabular}{cc}
        \toprule
        \textbf{Parameter} & \textbf{Value} \\
        \midrule
        Batch size & 256 \\
        Discount factor ($\gamma$) & 0.99 \\
        Actor learning rate & $3 \times 10^{-4}$ \\
        Critic learning rate & $3 \times 10^{-4}$ \\
        Target network update rate ($\lambda$) & $5 \times 10^{-3}$ \\
        Critic ensemble size ($K$) & 2 \\
        Critic target pessimistic coefficient ($\rho$) & 0.0 \\
        Rollouts collected before training & 10 \\
        Rollouts collected every 20k training steps & 10 \\
        \bottomrule
        \end{tabular}%
        }
        \caption{Parameters used in the real-world tasks for single-tasks policies.}
        \label{tab:real_world_rl_parameters}
    \end{minipage}\hfill
    \begin{minipage}[t]{0.45\textwidth}
        \centering
        \resizebox{\linewidth}{!}{%
        \begin{tabular}{l c c c c}
            \toprule
            & \multicolumn{4}{c}{OTQL} \\
            \cmidrule(lr){2-5} 
            Task & $\lambda$  & $w_{max}$ & $N$ & NFE\\
            \midrule
            \texttt{stack cups}        & 0.75& 10& 5 & 3\\
            \texttt{pen in penholder}       & 0.75& 10& 5 & 3\\
            \texttt{open door}  & 0.75& 10& 5 & 3\\
            \bottomrule
        \end{tabular}%
        }
        \vspace{0.1in}
        \caption{Hyperparameters used for OTQL in the real-world tasks.}
        \label{tab:real_world_training_hyperparameters}
    \end{minipage}
    
\end{table}

\begin{table}[htbp]
    \centering
    
    \begin{minipage}[c]{0.5\textwidth}
        \centering
        \resizebox{\linewidth}{!}{%
        \begin{tabular}{cc}
        \toprule
        \textbf{Parameter} & \textbf{Value} \\
        \midrule
        Batch size & 64 \\
        Discount factor ($\gamma$) & 0.99 \\
        Actor learning rate & $1 \times 10^{-4}$ \\
        Critic learning rate & $3 \times 10^{-4}$ \\
        Target network update rate ($\lambda$) & $5 \times 10^{-3}$ \\
        Critic ensemble size ($K$) & 2 \\
        Critic target pessimistic coefficient ($\rho$) & 0.0 \\
        \bottomrule
        \end{tabular}%
        }
        \caption{Parameters used in the real-world tasks for SmolVLA.} 
        \label{tab:real_world_rl_parameters_smolvla}
    \end{minipage}\hfill
    \begin{minipage}[c]{0.45\textwidth}
        \centering
        \resizebox{\linewidth}{!}{%
        \begin{tabular}{l c c c c}
        \toprule
        & \multicolumn{4}{c}{OTQL} \\
        \cmidrule(lr){2-5} 
        Task & $\lambda$  & $w_{max}$ & $N$ & NFE\\
        \midrule
        \texttt{pen in penholder}        & 1.0& 10& 1 & 3\\
        \texttt{move knight}       & 1.0& 10& 1 & 3\\
        \bottomrule
        \end{tabular}
        }
        \vspace{0.1in}
        \caption{Hyperparameters used for OTQL in the real-world tasks for SmolVLA adaptation.}
        \label{tab:real_world_training_hyperparameters_smolvla}
    \end{minipage}
    
\end{table}

\section{Hardware}
\label{sec:appendix_real_exp}
\begin{figure}[h]
    \vspace*{-1em}
    \centering
    \includegraphics[width=\linewidth]{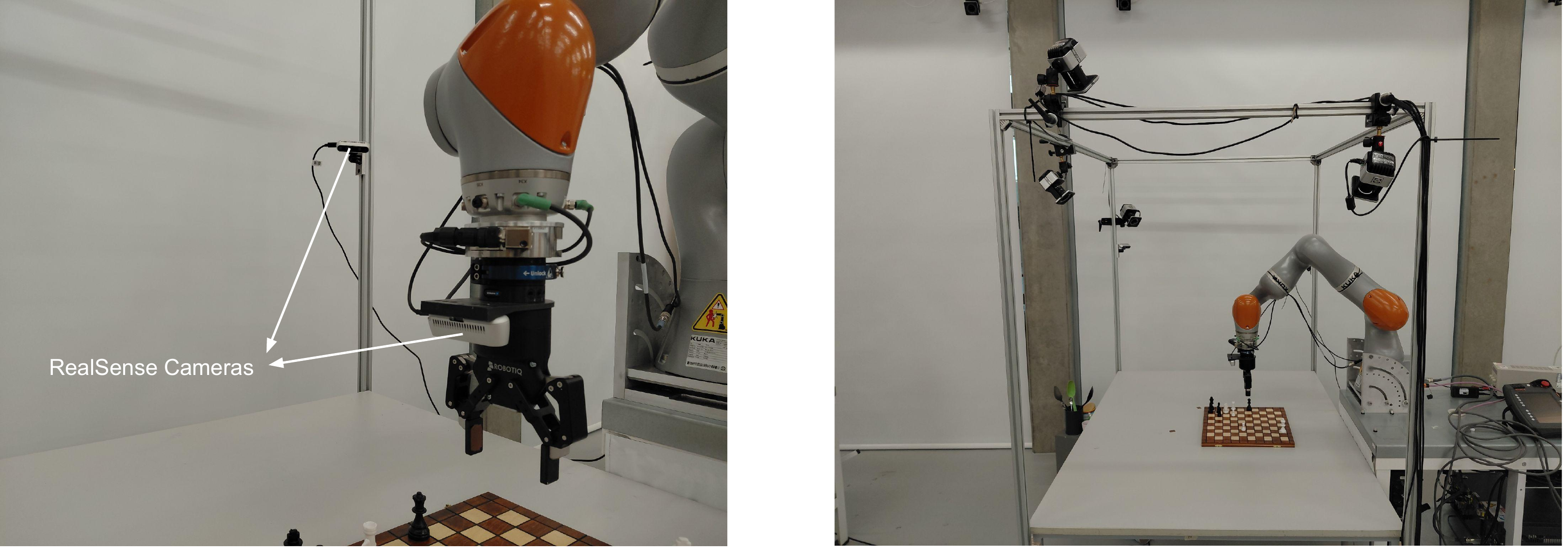}
    \caption{real-world robot setup}
    \label{fig:robot}
\end{figure}

\paragraph{Experiment setup.} The real-world experiments were performed on a KUKA IIWA 14 robot with a Robotiq 2F-85 gripper and we used 2 Realsense D415 cameras---one mounted on the end-effector and one providing an external view of the workspace. The entire setup can be seen in \Cref{fig:robot}. The real-world inference was performed using a desktop PC with (CPU, RAM, GPU): Intel(R) i9-9900KF, 64GB, NVIDIA GeForce RTX 2080. Data collection was performed using a 6D Spacemouse device.

\paragraph{Training and simulation evaluation.} Training and evaluation for the simulated tasks were performed using a workstation with (CPU, RAM, GPU): AMD Ryzen Threadripper PRO 5965WX 24-Cores, 128GB, $2\times$ NVIDIA GeForce RTX 4090.



\end{document}